# Detection and Analysis of Human Emotions through Voice and Speech Pattern Processing

Poorna Banerjee Dasgupta

*M.Tech Computer Science and Engineering, Nirma Institute of Technology*
*Ahmedabad, Gujarat, India*

***Abstract*** *— The ability to modulate vocal sounds and generate speech is one of the features which set humans apart from other living beings. The human voice can be characterized by several attributes such as pitch, timbre, loudness, and vocal tone. It has often been observed that humans express their emotions by varying different vocal attributes during speech generation. Hence, deduction of human emotions through voice and speech analysis has a practical plausibility and could potentially be beneficial for improving human conversational and persuasion skills. This paper presents an algorithmic approach for detection and analysis of human emotions with the help of voice and speech processing. The proposed approach has been developed with the objective of incorporation with futuristic artificial intelligence systems for improving human-computer interactions.*

## I. INTRODUCTION

The human voice consists of sounds produced by a human being using the vocal folds for carrying out acoustic activities such as talking, singing, laughing, shouting, etc. The human voice frequency is specifically a part of the human sound production mechanism in which the vocal cords or folds are the primary source of generated sounds. Other sound production mechanisms produced from the same general area of the body involve the production of unvoiced consonants, clicks, whistling and whispering. Generally, the mechanism for generating the human voice can be subdivided into three parts; the lungs, the vocal folds within the larynx, and the articulators [1].

The human voice and associated speech patterns can be characterized by a number of attributes, the primary ones being *pitch, loudness or sound pressure, timbre, and tone* [2].

*Pitch* is an auditory sensation in which a listener assigns musical tones to relative positions on a musical scale based primarily on their perception of the frequency of vibration [5][6]. Pitch can be quantified as a frequency, but it is based on the subjective perception of a sound wave. Sound oscillations can be measured to obtain a frequency in hertz or cycles per second. The pitch is independent of the intensity or amplitude of the sound wave. A high-pitched sound indicates rapid oscillations, whereas, a low-pitched sound corresponds to slower oscillations. Pitch of complex sounds such as speech and musical notes corresponds to the repetition rate of periodic or nearly-periodic sounds, or the reciprocal of the time interval between similar repeating events in the sound waveform.

*Loudness* is a subjective perception of sound pressure and can be defined as the attribute of auditory sensation, in terms of which, sounds can be ordered on a scale ranging from quiet to loud [7]. Sound pressure is the local pressure deviation from the ambient, average, or equilibrium atmospheric pressure, caused by a sound wave [9]. *Sound pressure level* (SPL) is a logarithmic measure of the effective pressure of a sound relative to a reference value and is often measured in units of decibel (dB). The lower limit of audibility is defined as SPL of 0 dB, but the upper limit is not as clearly defined.

*Timbre* is the perceived sound quality of a musical note, sound or tone [5]. Timbre distinguishes different types of sound production and enables listeners to distinguish different instruments in the same category. The physical characteristics of sound that determine the perception of timbre include spectrum and envelope. Figure 1 shows a sound wave with its temporal envelope marked in red. In simple terms, timbre is what makes a particular sound be perceived differently from another sound, even when they have the same pitch and loudness.

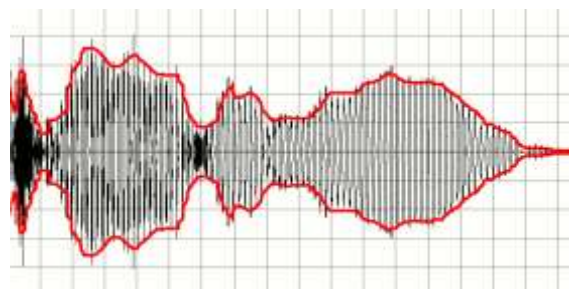

Fig 1: A sound wave's envelope marked in red [5]

*Tone* is the use of pitch in language to distinguish lexical or grammatical meaning – that is, to distinguish or to inflect words [8]. All verbal languages use pitch to express emotional and other paralinguistic information and to convey emphasis, contrast, and other such features.

It has been frequently observed that the tonal quality of the human voice changes while expressing various emotions [3][4]. With different emotions and moods, not only does the tonal quality vary, but the associated speech patterns change too. For instance, people may tend to talk in loud voices when angry and use shrill or high-pitched voices when in a scared or panicked emotional state. Some people tend to ramble when they get excited or nervous. On the contrary, when in a pensive emotional state, people tend to speak slowly and make longer pauses, thereby indicating an increase in time spacing between consecutive words of their speech.

Detection of human emotions through voice- and speech-pattern analysis can prove to be beneficial in improving conversational and persuasion skills, especially in those circumstances or applications where direct face-to-face human interaction is not possible or preferred. The practical applications of human emotion detection through voice and speech processing could be numerous. Some such plausible real world situations could be while conducting marketing surveys with customers through telephonic conversations, participating in anonymous online chat-rooms, conducting business voice-conferences and so on.

This paper presents an algorithmic approach which aids in detection of human emotions, by analyzing various voice attributes and speech patterns. Section II describes the proposed algorithmic approach in detail, along with the obtained results. Section III is devoted to result analysis. Finally, section IV elucidates the conclusions derived from the proposed study and future scope of work.

## II. ALGORITHMIC APPROACH TO DETECTION OF HUMAN EMOTIONS

This section describes an algorithmic approach for deducing human emotions through voice- and speech-pattern analysis. In order to achieve this objective, three test cases have been examined, corresponding to the three emotional states: *normal* emotional state, *angry* emotional state, and *panicked* emotional state. For carrying out the analysis, four vocal parameters have been taken into consideration: pitch, SPL, timbre, and time gaps between consecutive words of speech. In order to quantitatively represent timbre, its temporal envelope for advance and decay times has been considered. The primary function of the proposed algorithmic approach is to detect different emotional states by analyzing the deviations in the aforementioned four parameters from that of the normal emotional state. The proposed analysis was carried out with the help of software packages such as MATLAB and Wavepad.

- **Case 1**: *Normal* emotional state
  This test case involves statistics for pitch, SPL, timbre, and word-timing gaps derived from speech samples that were orated while the speaker was in a relaxed and normal emotional state. This test case serves as the basis for the remaining two test cases. All the parameter statistics indicate mean values derived from the speech samples. As shown in Table I, for the purpose of demonstration, statistics for two speech samples have been analyzed.

TABLE I
AVERAGE VALUES OF VOCAL STATISTICS OBTAINED FROM RECORDED SPEECH SAMPLES FOR A NORMAL EMOTIONAL STATE

|  | Pitch (Hz) | SPL (dB) | Timbre ascend time (s) | Timbre descend time (s) | Time gaps between words (s) |
|---|---|---|---|---|---|
| Speech Sample 1 | 1248 Hz | Gain -50 dB | 0.12 s | 0.11 s | 0.12 s |
| Speech Sample 2 | 1355 Hz | Gain -48 dB | 0.06 s | 0.05 s | 0.12 s |

- **Case 2**: *Angry* emotional state
  This test case involves statistics for pitch, SPL, timbre, and word-timing gaps derived from speech samples that were orated while the speaker was in an agitated emotional state, typically characterized by increased vocal loudness and pitch. All the parameter statistics indicate mean values derived from the speech samples, as shown in Table II. The same speech samples that were earlier used in Case 1 have been used in Case 2, but with a different intonation typical of an agitated or angry emotional state.

TABLE II
AVERAGE VALUES OF VOCAL STATISTICS OBTAINED FROM RECORDED SPEECH SAMPLES FOR AN ANGRY EMOTIONAL STATE

|  | Pitch (Hz) | SPL (dB) | Timbre ascend time (s) | Timbre descend time (s) | Time gaps between words (s) |
|---|---|---|---|---|---|
| Speech Sample 1 | 1541 Hz | Gain – 30 dB | 0.13 s | 0.10 s | 0.09 s |
| Speech Sample 2 | 1652 Hz | Gain – 29 dB | 0.06 s | 0.04 s | 0.10 s |

- **Case 3**: *Panicked* emotional state
  This test case involves statistics for pitch, SPL, timbre, and word-timing gaps derived from speech samples that were orated while the speaker was in a panicked or overwhelmed emotional state. Speech samples that were earlier used in Case 1 have been used in Case 3, but with a different intonation typical of a panicked emotional state, as shown in Table III.

TABLE III
AVERAGE VALUES OF VOCAL STATISTICS OBTAINED FROM RECORDED SPEECH SAMPLES FOR A PANICKED EMOTIONAL STATE

|  | Pitch (Hz) | SPL (dB) | Timbre ascend time (s) | Timbre descend time (s) | Time gaps between words (s) |
|---|---|---|---|---|---|
| **Speech Sample 1** | 1443 Hz | Gain – 46 dB | 0.13 s | 0.09 s | 0.13 s |
| **Speech Sample 2** | 1560 Hz | Gain – 44 dB | 0.07 s | 0.04 s | 0.14 s |

### III. RESULTS ANALYSIS

The speech samples described in the previous section were recorded with the help of headphones that have an offset gain of approximately -60 dB and the speech data was sampled at the rate of ten sample points per second. By comparing Tables I and II, it can be seen that when in an agitated state, a significant increase occurs in the mean SPL and pitch values, accompanied by a decrease in the time spacing between consecutive words. In simple terms, this would indicate faster talking in a shrill and louder voice.

By comparing Tables I and III, it can be seen that when in a nervous or panicked state, there is a significant increase in the mean values of pitch, time spacing between consecutive words, and increased timbre ascending time. In simple terms, this would indicate a shrill voice with longer, sharp pauses.

By comparing the data presented in Tables 1-3, it can be decisively concluded that with varying emotions, the tonal parameters accordingly change as well. Establishing value-ranges for the aforementioned various vocal parameters can help in quantitatively assessing the extent of deviation from the standard basis, which in this study is the *normal* emotional state. The findings of the previous section are subjective and will vary depending on how a person reacts to a particular emotional situation. However, from the point of practicality, this feature of subjectivity can be exploited, especially while developing customizable applications, such as smart-phone apps that are user oriented.

### IV. CONCLUSIONS

Deduction of human emotions through voice and speech analysis has a practical plausibility and could potentially be beneficial for improving human conversational and persuasion skills. This paper presents an algorithmic approach for detection and analysis of human emotions on the basis of voice and speech processing. Three test cases have been examined, corresponding to the three emotional states: normal emotional state, angry emotional state, and panicked emotional state. Each case demonstrates characteristic associated vocal features which can help in distinguishing the corresponding emotional state.

As future scope of work, more complex emotional states can be analyzed such as cheerful, unhappy, brooding, surprised, etc. While conducting the proposed study, speech data were sampled at the rate of ten sample points per second. In order to improve the precision of the mean attribute values, greater number of sampling points can be considered. Furthermore, in the proposed approach, only four attributes were analyzed for detecting various emotional states. To improve the accuracy of detection, other secondary vocal attributes could be assessed as well.

### REFERENCES

[1] Titze, I.R. *Principles of Voice Production*. Prentice Hall, ISBN 978-0-13-717893-3, 1994.
[2] Trevor R Agus, Clara Suied, Simon J Thorpe, Daniel Pressnitzer. *Characteristics of human voice processing*. IEEE International Symposium on Circuits and Systems (ISCAS), Paris, France. pp.509-512, May 2010.
[3] Smith, BL; Brown, BL; Strong, WJ; Rencher, AC. *Effects of speech rate on personality perception*. Language and speech. 18 (2): 145–52. PMID 1195957, 1975.
[4] Williams, CE; Stevens, KN. *Emotions and speech: some acoustical correlates.* The Journal of the Acoustical Society of America. Vol. 52 (4): 1238–50. PMID 4638039, 1972.
[5] Richard Lyon & Shihab Shamma. *Auditory Representation of Timbre and Pitch*. Harold L. Hawkins & Teresa A. McMullen. Auditory Computation. Springer. pp. 221–23. ISBN 978-0-387-97843-7, 1996.
[6] Plack, Christopher J.; Andrew J. Oxenham; Richard R. Fay, eds. *Pitch: Neural Coding and Perception*. Springer. ISBN 0-387-23472-1, 2001.
*[7]* American National Standards Institute. *American national psychoacoustical terminology*. S3.20, American Standards Association, 1973.
[8] R.L. Trask. *A Dictionary of Phonetics and Phonology*. Routledge, 2004.
[9] Morfey, Christopher L. *Dictionary of Acoustics*. San Diego Academic Press, ISBN 978-0125069403, 2001.

### AUTHOR'S PROFILE

**Poorna Banerjee Dasgupta** has received her B.Tech & M.Tech Degrees in Computer Science and Engineering from Nirma Institute of Technology, Ahmedabad, India. She did her M.Tech dissertation at Space Applications Center, ISRO, Ahmedabad, India and has also worked as Assistant Professor in Computer Engineering dept. at Gandhinagar Institute of Technology, Gandhinagar, India from 2013-2014 and has published several research papers in reputed international journals. Her research interests include image processing, high performance computing, parallel processing and wireless sensor networks.